# Lighting the Night with Generative Artificial Intelligence

Tingting Zhou[1]; Feng Zhang[2*]; Haoyang Fu[1]; Baoxiang Pan[3]; Renhe Zhang[2]; Feng Lu[4]; Zhixin Yang[1]

[1]College of Physics and Electronical Information Engineering, Zhejiang Normal University, Jinhua 321004, China

[2] Key Laboratory of Polar Atmosphere-Ocean-Ice System for Weather and Climate of Ministry of Education/ Shanghai Key Laboratory of Ocean-Land-Atmosphere Boundary Dynamics and Climate Change, Department of Atmospheric and Oceanic Sciences & Institutes of Atmospheric Sciences, Fudan University, Shanghai, China

[3]Institute of Atmospheric Physics, Chinese Academy of Sciences, Beijing, 100029, China

[4]CMA Key Laboratory for Cloud Physics, Weather Modification Center, China Meteorological Administration (CMA), Beijing, China

*Corresponding author(s). E-mail(s): fengzhang@fudan.edu.cn

## Abstract

The visible light reflectance data from geostationary satellites is crucial for meteorological observations and plays an important role in weather monitoring and forecasting. However, due to the lack of visible light at night, it’s impossible to conduct continuous all-day weather observations using visible light reflectance data. This study pioneers the use of generative diffusion models to address this limitation. Based on the multi-band thermal infrared brightness temperature data from the Advanced Geostationary Radiation Imager (AGRI) onboard the Fengyun-4B (FY4B) geostationary satellite, we developed a high-precision visible light reflectance generation model, called Reflectance Diffusion (RefDiff), which enables 0.47 μm, 0.65 μm, and 0.825 μm bands visible light reflectance generation at night. Compared to the classical models, RefDiff not only significantly improves accuracy through ensemble averaging but also provides uncertainty estimation. Specifically, the SSIM index of RefDiff can reach 0.90, with particularly significant improvements in areas with complex cloud structures and thick clouds. The model's nighttime generation capability was validated using VIIRS nighttime product, demonstrating comparable performance to its daytime counterpart. In summary, this research has made substantial progress in the ability to generate visible light reflectance at night, with the potential to expand the application of nighttime visible light data.

## 1. Introduction

The Sun serves as a primary source of energy and light for the Earth, with different materials exhibiting unique reflectance properties across various wavelengths of sunlight. During the day, humans can naturally observe and interpret the material world by detecting sunlight reflected off objects. However, as darkness descends, our visual capabilities are severely constrained. To surmount this limitation, human ingenuity led to the creation of torches, lamps, electric lights, and eventually advanced technologies like infrared and night vision equipment, progressively enhancing our ability to 'see' in the dark (Task, 2001). As technology has advanced, satellite remote sensing (American Meteorological Society, 2009) has emerged as a powerful tool for Earth observation, mimicking the role of the human eye, monitoring Earth both day and night.

Geostationary meteorological satellites can conduct high-frequency and wide-range monitoring of cloud evolution, playing a key role in the retrieval of meteorological elements, atmospheric process monitoring, and weather forecasting(Schmit et al., 2017). Orbiting satellites mainly include the U.S. GOES series (Goodman et al., 2019) the European Meteosat series (Stuhlmann et al., 2005), the Japanese Himawari series (Bessho et al., 2016), and Chinese Fengyun-4 series (Xian et al., 2021). The data provided by these satellites cover a spectral range from visible light to thermal infrared, with a visible light resolution of up to 500 meters, thermal infrared resolution ranging from 2 to 4 kilometers, and a temporal resolution of approximately 10 to 15 minutes. Geostationary satellite data typically includes multiple visible and thermal infrared channels. Among these, the red, green, and blue visible light bands can be combined to generate true-color images. These images align more closely with human visual perception, facilitating a more intuitive observation of cloud evolution. Furthermore, studies have shown that visible light data often outperform thermal infrared data in capturing the fine structural details of extreme weather events, such as the eyewall structure of tropical cyclones (TC) and the organized spiral rainbands(Velden et al., 2006; Yao et al., 2024). Due to the absence of visible light at night, cloud observation during this time depends solely on thermal infrared data (Min et al., 2020; Zhao et al., 2023). Therefore, developing an effective method for generating visible light at night by converting thermal infrared and other auxiliary data could obtain continuous all-day visible light data. This would provide valuable support for weather system monitoring using visible light.

Reflectance, as an intrinsic physical property of objects, is primarily determined by their material composition and surface structure and remains unaffected by variations in incident light conditions. Based on this principle and atmospheric radiative transfer theory, the radiative characteristics of objects in both thermal infrared and visible light channels are inherently related. This correlation provides a theoretical basis for employing machine learning techniques (Yuan et al., 2020) to construct nonlinear transformation models between thermal infrared and visible light data, thereby enabling nighttime visible light remote sensing. Kim et al. (2019) initially applied deep learning method to generate the 0.675 μm channel by utilizing data from the 10.8 μm channel of the Communication, Ocean, and Meteorological Satellite/Meteorological Imager (COMS/MI). Subsequently, several researchers (B. Chen et al., 2021; Han et al., 2022; Kim et al., 2020) explored nighttime visible light generation by employing various thermal infrared channels, auxiliary data, and deep learning techniques, primarily UNet (Ronneberger et al., 2015) and generative adversarial networks (GANs) (Goodfellow et al., 2020). These studies not only confirmed the feasibility of generating visible light data from thermal infrared data but also demonstrated the capability of the deep learning methods to produce high-quality nighttime visible light data. However, traditional deep learning approaches, such as UNet, typically optimize model parameters by minimizing loss functions like root mean square error (RMSE) or mean absolute error (MAE), providing a single output that often fails to capture extreme values (Chollet, 2016; Zhou et al., 2018). Methods based on GAN also face the challenge of generating extreme values (Dhariwal and Nichol, 2021), and due to the differences in data distribution between day and night, GAN may also suffer from domain shift (Zhang et al., 2019).

Developing an objective function that accurately reflects the “similarity” between model outputs and targets is crucial for improving model performance. Unlike traditional deep learning methods, probabilistic diffusion models adopt likelihood-based generative approaches, with training objectives focused on enhancing the model’s probabilistic estimation of observed data. This enables the model to better capture the features of target distributions (Ho et al., 2020). A key advantage of diffusion models is their progressive generation process, allowing output adjustments based on initial conditions (Ho and Salimans, 2022; Zhang et al., 2023). This approach not only enhances the precision of generated results but also ensures comprehensive coverage of all potential outputs. Probabilistic diffusion models are particularly suited for scenarios requiring consideration of multiple potential outcomes while providing accurate predictions under specific conditions (H. Xiao et al., 2024). Studies have demonstrated that these models exhibit unique advantages in handling complex systems such as short-term weather forecasting (Gao et al., 2023), remote sensing data downscaling (Y. Xiao et al., 2024), and meteorological data assimilation (Huang et al., 2024).

This study proposes a generative diffusion model, RefDiff, which establishes a mapping relationship between geostationary satellite thermal infrared and visible light channels. Compared to traditional deep learning models, RefDiff effectively captures the distribution of visible light reflectance and uses thermal infrared channels to constrain the corresponding visible light reflectance distribution. This enables the transformation of thermal infrared data into visible light information, generating nighttime visible light reflectance in three bands: 0.47 µm, 0.65 µm, and 0.825 µm.

# 2. Data

## 2.1 Data Sources

### *2.1.1 Satellite Data from FY4B/AGRI*

The data utilized in this study were obtained from the Advanced Geostationary Radiation Imager (AGRI) onboard the Fengyun-4B (FY4B) geostationary satellite. FY4B, the second satellite in China's Fengyun-4 geostationary meteorological series, was successfully launched on June 3, 2021. It commenced providing observation data and application services on June 1, 2022. On March 5, 2024, FY4B replaced FY4A in its operational role, continuing to provide services in geostationary orbit at 105°E. Due to insufficient data accumulation after the orbital change, this study selected data collected between June 1, 2022, and January 31, 2024. The training set comprised data from June 2022 to May 2023, while the testing set consisted of data from June 2023 to January 2024.

FY4B/AGRI features 14 observation channels, encompassing two visible light, one near-infrared, three short-wave infrared, two mid-wave infrared, three water vapor, and four long-wave infrared channels. It covers a full-disk observation range from 52°E to 148°W longitude and 81°S to 81°N latitude. For RGB composite imagery generation, this study targets the generation of 0.47µm, 0.65µm, and 0.825µm bands. The constructed model employs AGRI's three water vapor channels (6.25µm, 6.96µm, and 7.42µm) and four long-wave infrared channels (8.55µm, 10.8µm, 12µm, and 13.3µm) as input data. Notably, the three short-wave infrared channels (1.379µm, 1.61µm, and 2.25µm) were excluded due to their susceptibility to sunlight interference. Similarly, the two mid-wave infrared channels (3.75µm at 2 km and 4 km resolutions) were omitted due to their sensitivity to sunlight during twilight periods. Given the significant impact of satellite zenith angle (SAZ) variation on data quality, SAZ was incorporated as a crucial input variable to enhance the model's accuracy and adaptability.

#### *2.1.2 Land Cover Classification Data*

To account for the impact of surface types on radiation budget and energy balance, this study incorporates high-precision land cover classification data into the model construction. The Land Cover Classification Gridded Maps, provided by the European Space Agency (ESA) Climate Change Initiative (CCI), were utilized. This dataset offers detailed surface cover information at a high spatial resolution of 500 meters and is accessible through the Copernicus Climate Data Store (https://cds.climate.copernicus.eu/cdsapp#!/dataset/satellite-land-cover?tab=overview). Given the relatively slow interannual changes in land cover types and to maintain data consistency while simplifying the processing workflow, this study uniformly employed the 2022 land cover data.

#### *2.1.3 VIIRS Panchromatic Day-Night band (DNB) Calibrated Radiance Product*

To obtain visible light at night, the Visible Infrared Imaging Radiometer Suite (VIIRS) on the Suomi National Polar Orbit Partnership (Suomi NPP) satellite uses moonlight reflection for night Earth observation (Chen et al., 2021; Elvidge et al., 2017). To evaluate the performance of nighttime visible light generation, this study employed the data from the VIIRS Panchromatic Day-Night Band (DNB) Calibrated Radiance Product (VNP02DNB) to assess the nighttime TC data generated by the proposed model. VNP02DNB, an L1B product from the Suomi National Polar-orbiting Partnership (SNPP) platform, is derived from calibrated radiance measurements of the Day-Night Band (DNB) of the Visible Infrared Imaging Radiometer Suite (VIIRS) sensor. The DNB spectral range encompasses 0.5μm to 0.9μm, capturing a broad spectrum of information from visible to near-infrared wavelengths, with a spatial resolution of 750 meters. The data were obtained from the Level-1 and Atmosphere Archive & Distribution System Distributed Active Archive Center (LAADS DAAC).

### 2.2 Data Processing

#### *2.2.1 Data Preprocessing*

This study constructed a comprehensive dataset using multi-source remote sensing data. The primary data sources include visible, near-infrared, and multiple thermal infrared bands from FY4B/AGRI, complemented by auxiliary data such as satellite zenith angle (SAZ), land cover classification, solar zenith angle (SOZ), and nighttime light data from VNP02DNB. SOZ is utilized to distinguish between day and night data, while VNP02DNB is employed to evaluate the generated nighttime visible light. Due to the varying spatial and temporal characteristics of each data source, unified resampling and spatiotemporal alignment were performed during preprocessing. Using AGRI's full-disk 4km resolution L1 product as the reference, all datasets were resampled to a 4km resolution latitude-longitude grid. To ensure data quality, the study area was confined to the AGRI coverage, spanning from 68°E to 198°E and 65°S to 65°N, corresponding to a $3252 \times 3525$ pixel grid. Both the training and testing sets were selected from this region, excluding areas with excessively large SAZ to minimize potential biases.

#### *2.2.2 Dataset Construction*

To construct an efficient and accurate dataset, this study implemented rigorous filtering and quality control measures for the FY4B/AGRI data. Given that AGRI does not provide nighttime visible light data, only daytime data were utilized for model training. To ensure data quality, samples with solar zenith angles (SOZ) less than 85° were selected, effectively mitigating abnormally low visible light values caused by low solar elevation angles. This approach enhances the data's reliability and representativeness. Considering AGRI's service commencement on June 1, 2022, data from June 1, 2022, to May 31, 2023, were used for training, while data from June 2023 to January 2024 were employed for testing. To minimize data redundancy and optimize the training set size, data were sampled at two-hour intervals. Samples with invalid data were excluded, and the remaining data were segmented into $256 \times 256$ pixel grid, yielding 262,947 valid training samples. The independent test set comprises data from 4:00 UTC on the 1st, 11th, and 21st of each month between June 2023 and January 2024. Furthermore, as the model primarily aims to generate nighttime visible light data, an additional independent test dataset focusing on several nighttime TC cases was established, utilizing processed VNP02DNB data to evaluate the generation performance of nighttime visible light.

## 3. Methods

## 3.1 Model

#### *3.1.1 Diffusion Model*

Diffusion Models are generative models that simulate a thermodynamic-inspired process by gradually adding noise to data (forward process) and denoising it (reverse process) through a neural network to approximate the target distribution (Ho et al., 2020).

In the forward process, Gaussian noise is incrementally added to the visible light data, transforming it into standard Gaussian noise. This transformation follows a series of Markov chains:

$$q(R_t|R_{t-1}) = \mathrm{N}\left(X_t;\ \sqrt{1-\beta_t}R_{t-1},\ \beta_t \mathrm{I}\right) \tag{1}$$

here, $R_t$ is the state of reflectance of visible light at time step *t*, $\beta_t$ is a predefined diffusion coefficient (usually between 0 and 1, and gradually increasing), and $1 \leq \mathrm{t} \leq \mathrm{T}$. When T is sufficiently large, $q(R_T|R_0)$ will approach a Gaussian distribution.

The reverse process aims to recover the original data from the noisy version. Estimating $q(R_{t-1}|R_t)$ is complex, so it is approximated by training a neural network $p_\theta$ to model the reverse diffusion:

$$p_\theta(R_{t-1}|R_t) = \mathrm{N}\left(R_{t-1};\ \mu(R_t,t), \textstyle\sum_\theta(R_t,t)\right) \tag{2}$$

Ho et al.(2020) proposed a simplified loss function to enhance sample quality and simplify implementation:

$$L_{simple}(\theta) \coloneqq \mathbb{E}_{t,X_0,\epsilon}[\| \epsilon - \epsilon_\theta(R_t,t) \|] \tag{3}$$

where $\bar{\alpha}_t = \prod_{s=1}^{t}\alpha_s$, $\epsilon \sim \mathcal{N}(0,\mathrm{I})$, $\epsilon_\theta$ is a function approximator intended to predict $\epsilon$ from $R_t$.

During inference, the process starts from standard Gaussian noise $z \sim \mathcal{N}(0,\mathrm{I})$, and the learned reverse diffusion process is applied iteratively:

$$R_{t-1} = \frac{1}{\sqrt{\alpha_t}}\left(R_t - \frac{1-\alpha_t}{\sqrt{1-\bar{\alpha}_t}}\epsilon_\theta(R_t,t)\right) + \sigma_t z \tag{4}$$

where $\alpha_t = 1 - \beta_t$, $\sigma_t^2 = \beta_t$.

Conditional diffusion models extend standard diffusion models by incorporating conditional information to generate samples that meet specific requirements (Dhariwal and Nichol, 2021; Ho and Salimans, 2022; Rombach et al., 2022). In this study, we utilize multiple channels of thermal infrared brightness temperature data from FY4B/AGRI, land use data, and SAZ data as conditional information *y*. The target variable $R$ represents reflectance of visible light information across three spectral bands. With the introduction of conditional information *y*, each step of the reverse diffusion process, i.e., estimating $R_{t-1}$ from $R_t$ as expressed by the neural network in Equation (2), becomes:

$$p_\theta(R_{t-1}|R_t,y) = \mathrm{N}\left(R_{t-1};\ \mu(R_t,t,y), \textstyle\sum_\theta(R_t,t,y)\right) \tag{5}$$

Accordingly, after introducing the condition $y$, the objective function becomes:

$$L_{cond}(\theta) = \mathbb{E}_{t,X_0,\epsilon}[\| \epsilon - \epsilon_\theta(R_t, t, y) \|] \quad (6)$$

### *3.1.2 Backbone*

For the reverse diffusion process, we employ a UNet architecture, enhanced by a multi-head attention mechanism and a residual network structure. The multi-head attention (Cordonnier et al., 2021) mechanism enables the model to focus on different spatial locations and channels, improving its ability to understand and generate complex patterns. The residual network (Zhang et al., 2018) aids in mitigating the vanishing gradient problem, facilitating better convergence and performance. The reverse process, utilizing deep learning, gradually removes the noise in successive steps. The input to the deep learning model consists of Gaussian noise and conditions, which include multi-channel brightness temperature, land cover, and satellite zenith angle. Figure 1 is the schematic representation of the methodology employed in this study. The first row reflects the sources of visible light and thermal infrared information, the second row represents the entire bandwidth of FY4/AGRI, and third row represents the model proposed in this study. For RefDiff, the forward process progressively adds noise to the visible light data until it becomes Gaussian noise, while the reverse process employs deep learning to iteratively denoise the data. The input to the deep learning model consists of Gaussian noise and conditional information, which includes multi-channel brightness temperature, land cover, and satellite zenith angle.

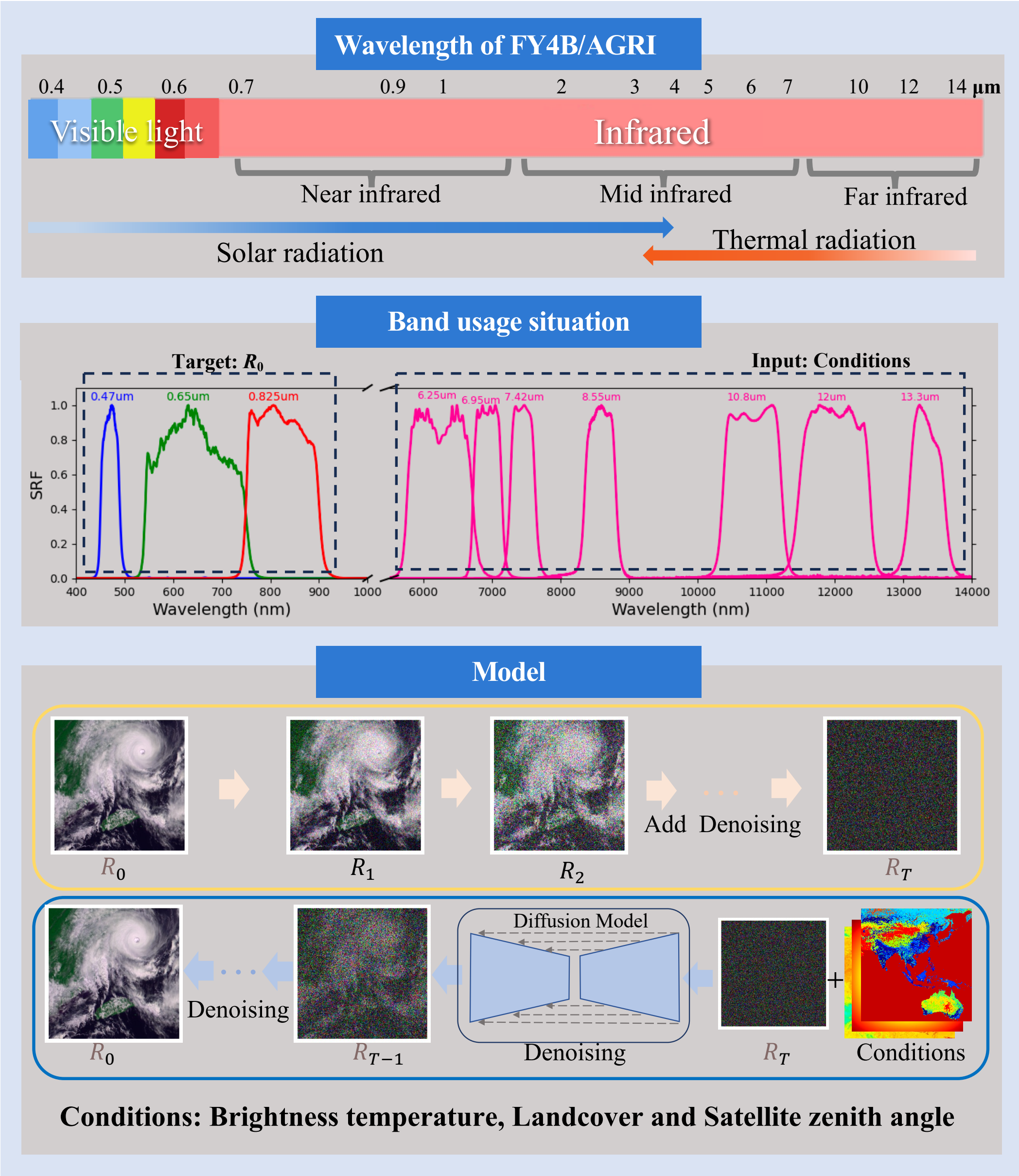

**Figure 1. Schematic diagram of the study.**

## 3.2 Evaluation Metrics

This study employs four metrics to assess the quality of the generated data: Mean Absolute Error (MAE), Root Mean Square Error (RMSE), Structural Similarity Index (SSIM) (Brunet et al., 2012; Wang et al., 2004), and Peak Signal-to-Noise Ratio (PSNR) (Korhonen and You, 2012). MAE calculates the average absolute difference between the pixel values of the generated data and the ground truth, with lower values indicating higher accuracy. RMSE measures the square root of the mean squared differences between the generated data and the ground truth, where smaller values reflect better alignment with the target data. SSIM evaluates image similarity by accounting for luminance, contrast, and structure, with a range of -1 to 1, where values closer to 1 signify greater similarity to the ground truth. PSNR, expressed in decibels (dB), is commonly used to assess image quality, with higher values representing superior quality.

## 4. Results and Discussion

### 4.1 Evaluation of RefDiff Performance

RefDiff can bring the ensemble average results closer to the target distribution by generating multiple ensemble members through probabilistic distribution sampling. To fully evaluate the performance of RefDiff, we compared its performance with that of classical deep learning models, and analyzed the impact of the number of ensemble members on performance (Li et al., 2024). To prevent the artificial inflation of performance by using data temporally close to the training set for testing, data from the entire study region were selected on the 1st, 11th, and 21st of each month between June 2023 and January 2024 for model testing and evaluation. The evaluation results are shown in Figure 2. Considering that the UNet (Ronneberger et al., 2015) and CGAN (Mirza and Osindero, 2014) architectures are among the most widely used deep learning models, the results of these two models were compared with those of RefDiff. All models were trained on the same dataset for 600 epochs, and their performance was evaluated across four metrics: MAE, RMSE, SSIM, and PSNR. Figures 2a-d present the performance of RefDiff with ensemble number of 1, 5, 10, 20, and 30, as well as the performance of UNet and CGAN. From the results above, it can be observed that the performance of RefDiff in all four metrics increases as the number of ensemble members grows. Moreover, RefDiff consistently outperforms both UNet and CGAN across all metrics, even when the ensemble number is just 1. These results indicate that RefDiff offers a distinct advantage over classical deep learning models such as UNet and CGAN in the task of visible light reflectance generation, and that its performance can be further enhanced through ensemble mean.

To objectively evaluate the effect of ensemble mean on model performance, we further analyzed the relationship between the number of ensemble members and various evaluation metrics of all the test data. As shown in Figure 2e, RMSE decreases significantly while SSIM increases substantially as the number of ensemble members grows. Figure 2f presents the proportion of true values that fall within the distribution of ensemble members, referred to as the Ensemble Coverage Proportion. The results show that as the number of ensemble members increases, the Ensemble Coverage Proportion increases rapidly. Both Figures 2e and 2f indicate that ensemble mean significantly improves the visible light generation accuracy and brings the generation results closer to the target distribution. At the same time, it can be seen that the performance stabilizes when the ensemble number reaches approximately 25. These trends suggest that, compared to deterministic methods, ensemble mean is more suitable for visible light generation tasks, providing more accurate and stable results. Furthermore, once the number of ensemble members exceeds 25, the improvement in evaluation metrics becomes negligible. Considering the computational cost of increasing the ensemble size and the accuracy requirements, we ultimately determined the number of ensemble members to be 30.

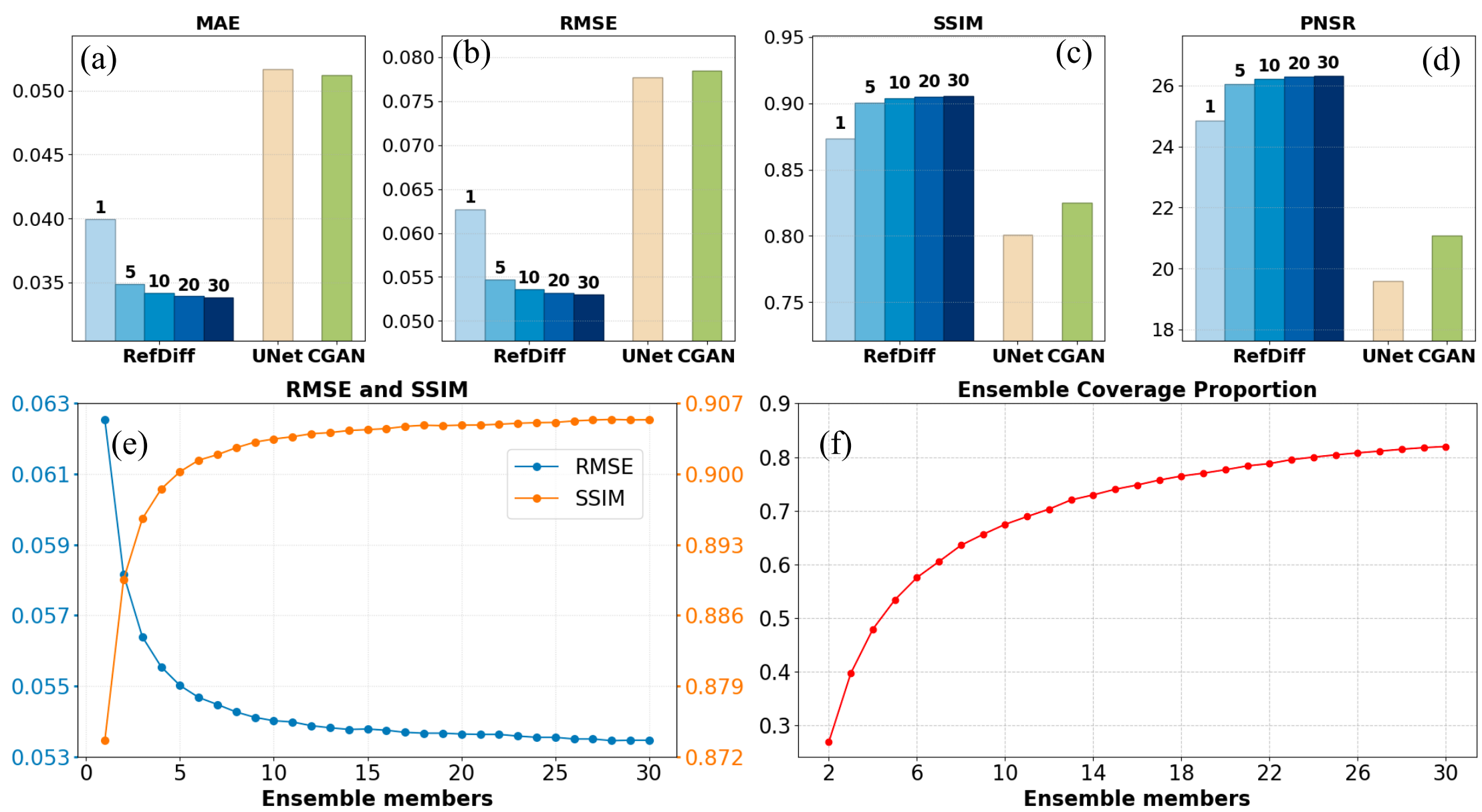

**Figure 2. Performance comparison of different models and the effect of the number of ensemble members. (a)-(d) show the comparison between RefDiff, UNet, and CGAN, with the numbers in RefDiff indicating the accuracy at ensemble numbers. (e) shows how RMSE and SSIM change with the number of ensemble members. (f) presents the proportion of true values falling between the minimum and maximum values of the ensemble members as the number of ensemble members changes.**

For a more intuitive comparison of the performance of the inversion at each wavelength, the results of three wavelengths were evaluated summarized in Table 1. The data shows that CGAN performs similarly to UNet in terms of MAE (~0.5) and RMSE (~0.7), but exhibits a clear advantage in SSIM, improving from approximately 0.8 to 0.82, and in PSNR, with an increase from 19 to 21. The RefDiff model, however, consistently outperforms both UNet and CGAN across all metrics. Specifically, compared to UNet and CGAN, RefDiff improves the SSIM from 0.80 and 0.82 to 0.91, and the PSNR from 19.58 and 21.09 to 25.39, respectively.

It is important to note that, compared to MAE, RMSE, and PSNR, SSIM not only considers pixel-level errors but also takes into account the image's structure, luminance, and contrast information, making it more suitable for evaluating image quality. In the field of image processing (Wang et al., 2003; Wang and Bovik, 2006), when SSIM is greater than 0.9, it is considered that the generated image is very close to the label image, and the difference is hardly noticeable to the human eye. When the SSIM value falls between 0.8 and 0.9, differences between the images become evident, particularly in terms of details and edges. However, when the SSIM value drops below 0.8, the image is considered to exhibit noticeable distortion or a degradation in quality. As shown in Table 1, both UNet and CGAN have SSIM values around 0.8, indicating that the differences between the generated images and the ground truth are easily observable. In contrast, RefDiff achieves an SSIM around 0.9, indicating a significant improvement in image quality compared to UNet and CGAN. These results clearly demonstrate RefDiff's superior ability in capturing fine structural features, showing that, in visible light generation tasks, generative diffusion models have a distinct advantage over traditional deep learning models.

**Table 1 Performance of the three models on MAE, RMSE, SSIM, and PSNR of daytime.**

| Band | Model | MAE | RMSE | SSIM | PSNR |
|---|---|---|---|---|---|
| 0.47μm | UNet | 0.0501 | 0.0757 | 0.803 | 19.18 |
| | CGAN | 0.0501 | 0.0752 | 0.819 | 21.05 |
| | RefDiff | **0.0329** | **0.0514** | **0.898** | **26.85** |
| 0.65μm | UNet | 0.0512 | 0.0784 | 0.798 | 19.98 |
| | CGAN | 0.0532 | 0.0813 | 0.827 | 21.13 |
| | RefDiff | **0.0339** | **0.0538** | **0.907** | **26.51** |
| 0.825μm | UNet | 0.0538 | 0.0791 | 0.803 | 19.59 |
| | CGAN | 0.0524 | 0.0790 | 0.831 | 21.10 |
| | RefDiff | **0.0344** | **0.0538** | **0.906** | **25.39** |

## 4.2 The Performance in TC Case

### *4.2.1 The performance of TC case on different models*

To effectively demonstrate the performance of each model, the typhoon Saola from 4:00 UTC on September 1, 2023, was selected for analysis. The results are presented in Figure 3. Rows 1 to 3 display the results for 0.47 μm, 0.65 μm, and 0.825 μm, respectively. The leftmost four columns show the AGRI values alongside the results generated by each model, while the rightmost three columns depict the deviations from the AGRI data. As shown in the left four columns, the high-reflectance regions corresponding to the TC are significantly underestimated by UNet and CGAN compared to RefDiff. Overall, the RefDiff model demonstrates clear superiority in both quantitative metrics and qualitative analysis, particularly excelling at capturing the intricate structures of complex weather systems. The fourth row of Figure 3 presents the RGB composite images from AGRI and the three models, offering a clear depiction of the TC's eye, the integrity of the eyewall, and the organization of the outer spiral rainbands. From the fourth row of Figure 3, it can be seen that compared to the results presented by AGRI, UNet performs the worst, with no information on the TC eye and minimal texture details of the eyewall and spiral rainbands. The results of the CGAN model slightly outperform those of UNet, as it partially captures the texture structure of the spiral rainbands. However, like UNet, it fails to generate information about the TC eye. RefDiff not only clearly captures the typhoon eye information but also provides a complete texture structure.

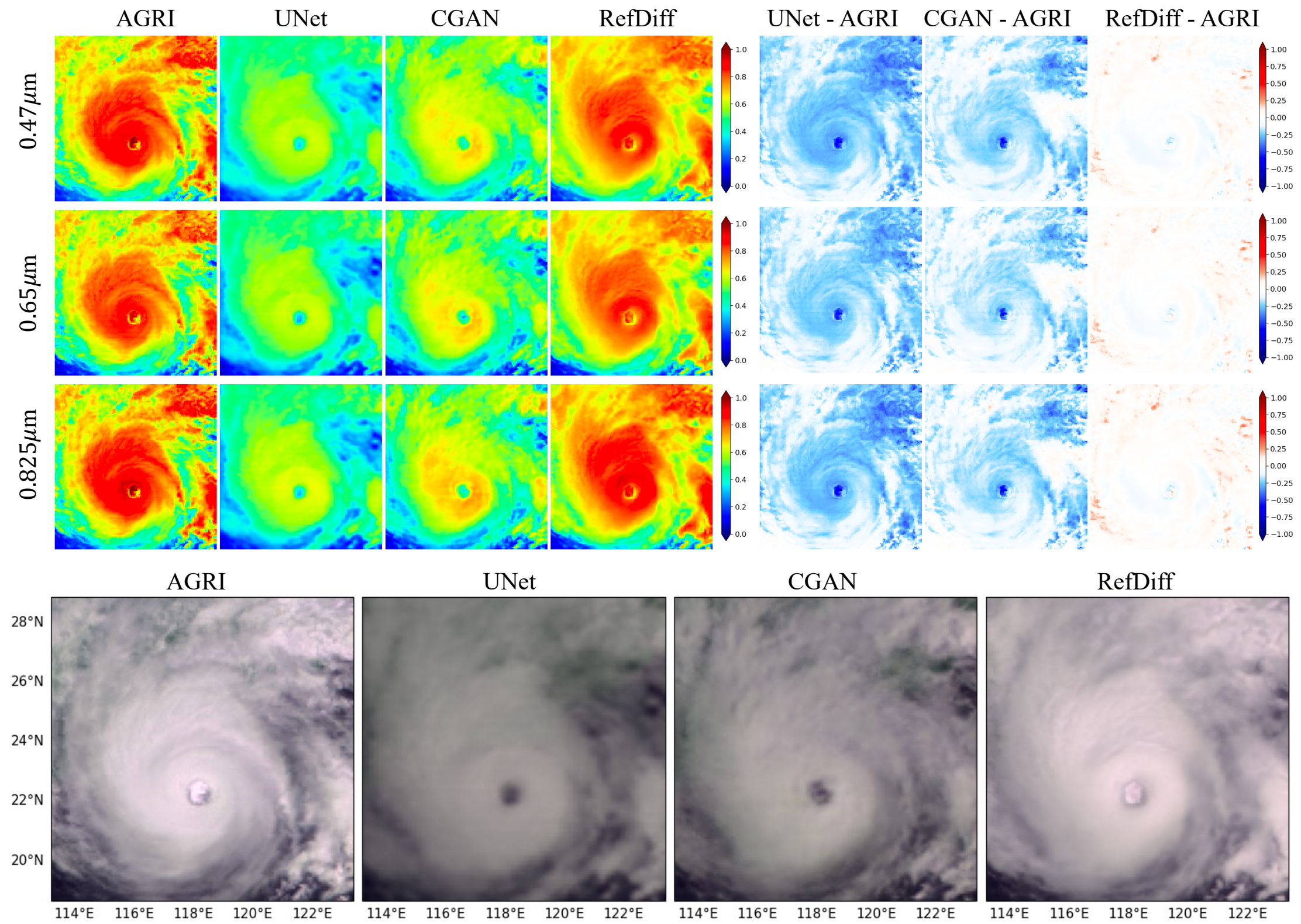

**Figure 3. TC case on September 1, 2023, at 04:00 UTC. Rows 1 to 3 display the FY4B/AGRI observational data at the wavelengths of 0.47 µm, 0.65 µm, and 0.825 µm, along with the results generated by UNet, CGAN, and RefDiff, as well as the differences between each model's results and the AGRI data. Row 4 presents the RGB composite true-color images for the three wavelengths.**

#### *4.2.2 The effect of ensemble member number on TC case*

To more intuitively observe the differences between different ensemble members, Figure 4 presents the results of 30 ensemble members for the TC case. From the figure, it can be observed that although the individual ensemble members are generally similar, there are still certain differences between them. The rightmost column of Figure 4 shows the ground truth from AGRI, where regions marked with black '×' indicate positions where the ground truth do not fall within the distribution of the ensemble members. The proportion of black '×' for all three bands is below 10%. Based on the position of the black '×', it can be seen that deviation are more likely to occur in the thick cloud regions surrounding the TC eye (with higher reflectance), while the ground truth in cloud-free or thin cloud region is typically covered by the distribution of multiple ensemble members.

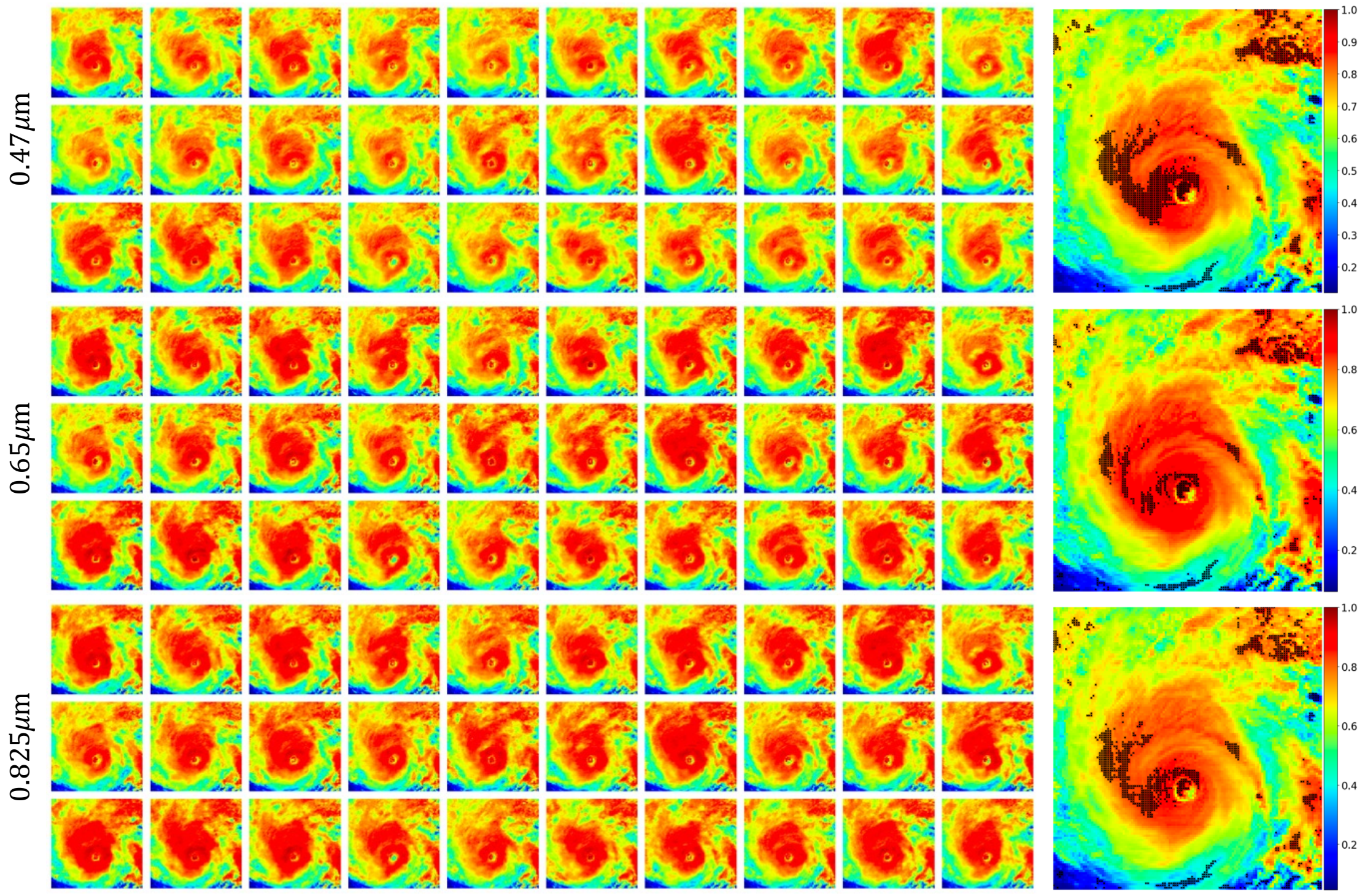

**Figure 4. 30 ensemble members generated by RefDiff and the ground truth from AGRI. The first to third rows display results for wavelengths of 0.47 μm, 0.65 μm, and 0.825 μm, respectively. The small images on the left show the results of the 30 ensemble members, while the large images on the right show the ground truth from AGRI. Black '×' on the ground truth images indicate pixels where the ground truth do not fall within the range of the ensemble member values.**

Figure 5 illustrates the distribution of individual ensemble members compared to the ensemble mean. The first column presents the ground truth from AGRI, the second shows the result of a single ensemble member, the third displays the ensemble mean derived from 30 members, and the fourth represents the standard deviation (Std) of individual results relative to the ensemble mean. The Std quantifies the uncertainty among ensemble members. Comparing the first and second columns, we observe that the output from a single ensemble member significantly deviates from the ground truth, indicating a tendency toward overestimation or underestimation. However, the ensemble mean effectively corrects the bias introduced by individual results. This demonstrates that RefDiff, through multiple iterations, can generate ensemble members that include more extreme values, enabling the ensemble mean distribution to better approximate the target distribution. In the fourth column, the Std of the 30 ensemble members reflects the uncertainty of the ensemble mean. It can be observed that, except for a few pixels in the TC region with relatively large Std (~0.125), the Std across most areas, including cloudy regions, remains within 0.05. This indicates that the generation capability of RefDiff is generally stable. In summary, this ensemble prediction method not only improves the overall accuracy of the model but also provides uncertainty estimation results, helping to better understand the model's performance and limitations in different scenarios.

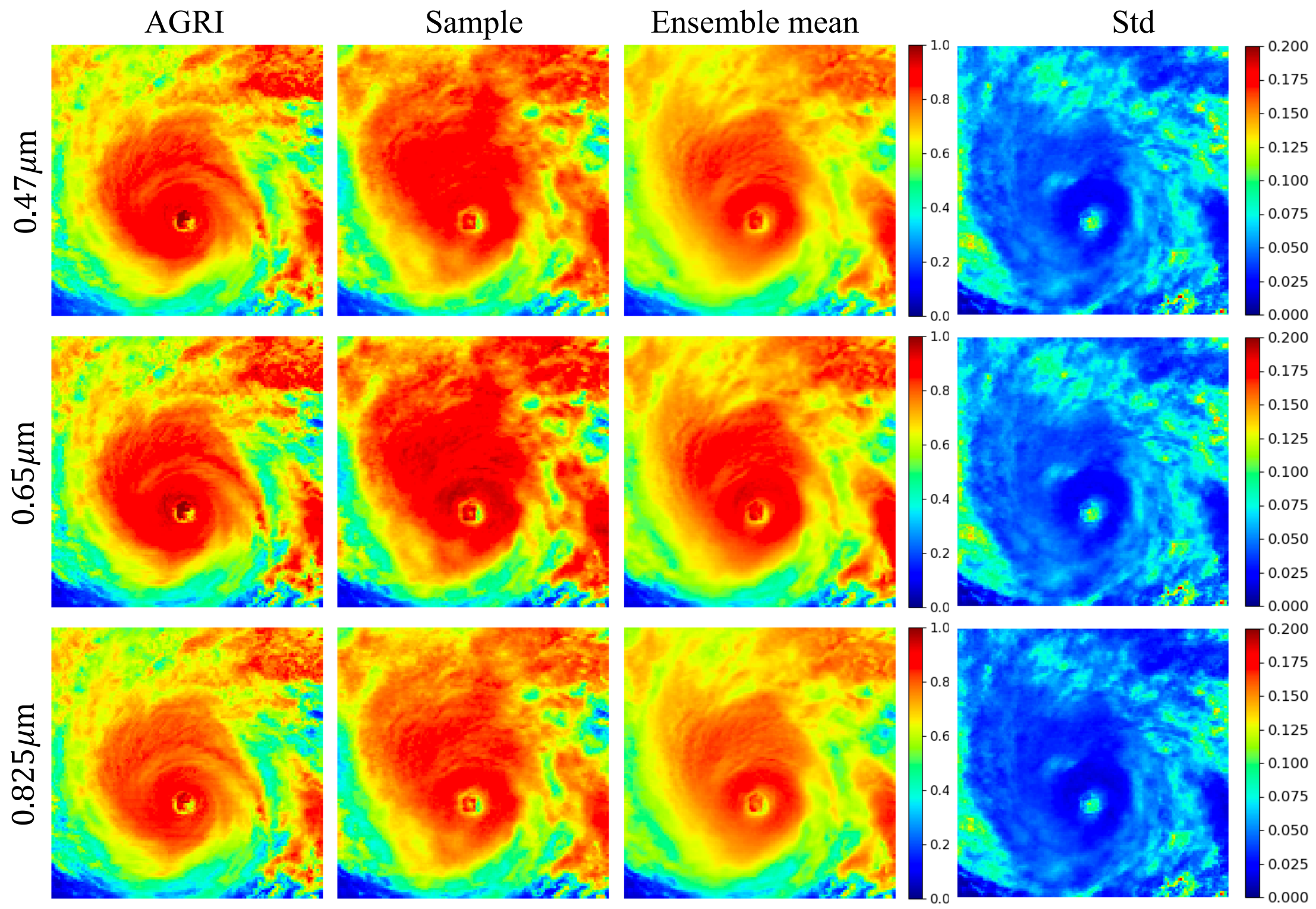

**Figure 5. The first column presents the labels from FY4B, the second column shows the results from a single ensemble member, the third column displays the ensemble mean of 30 members, and the fourth column illustrates the Std (uncertainty) of the 30 individual members from the ensemble mean.**

#### *4.2.3 Results of the entire study region*

In addition to quantitative assessments, this study visually presents the generation results for the entire study area during the landfall of Typhoon Doksuri on July 28, 2023. Figure 6a shows the situation at 10:00 UTC, with the TC located directly on the solar terminator. Under these lighting conditions, visible light reflectance is typically very low, making it difficult for RGB composite images to clearly display the TC's structure. However, the results produced by the RefDiff (Figure 6b) reveal the textural structure of the typhoon distinctly, demonstrating that the model can effectively generate visible light and has good consistency on the terminator line. This result strongly affirms the RefDiff model's capability to generate high-quality visible light data under all-day conditions. By leveraging the reflectance generated by the RefDiff model, stable and continuous observations of extreme weather events such as TC and strong convection can be based on visible light.

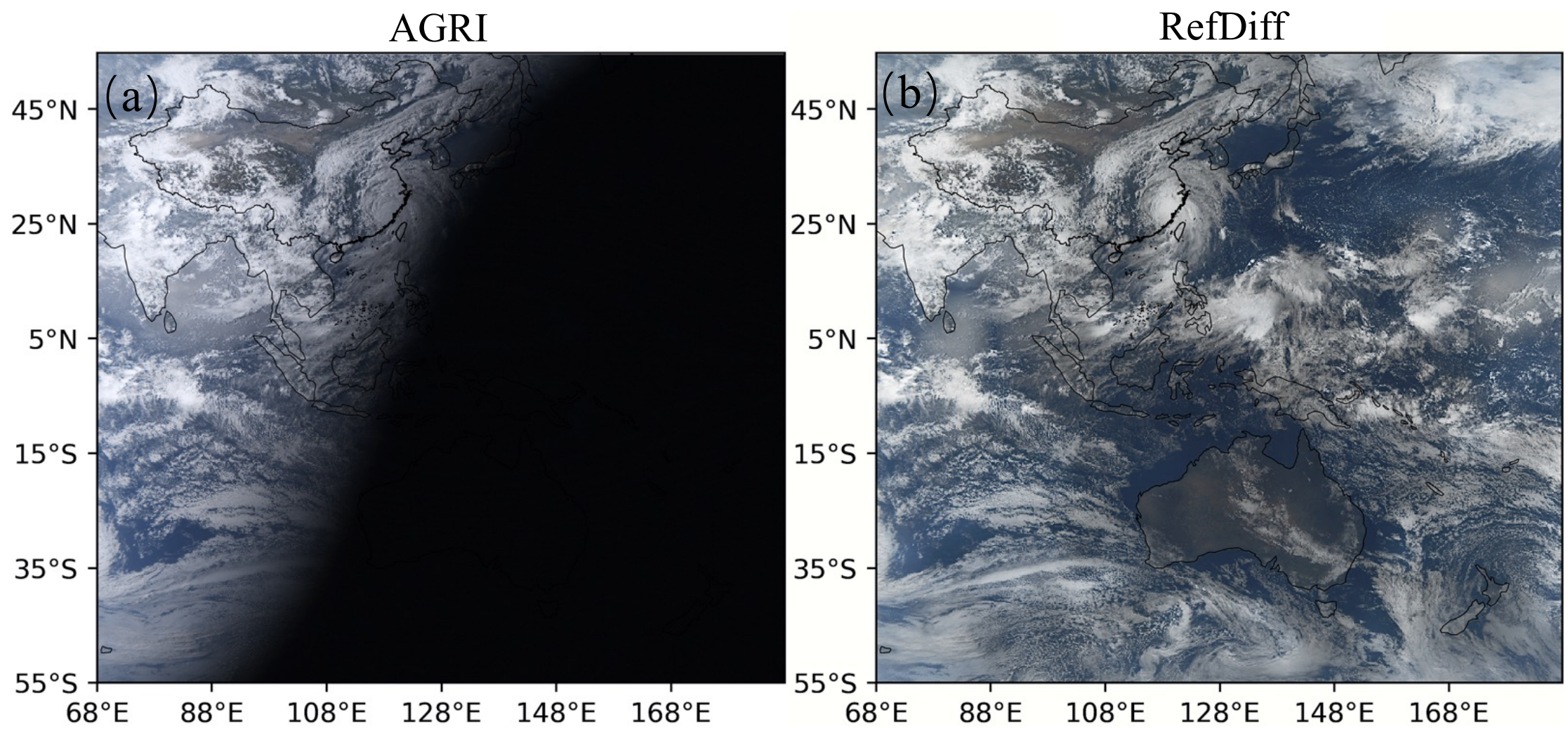

**Figure 6. TC case on July 28, 2023, at 10:00 UTC, where (a) represents the AGRI data, and (b) shows the results generated by the RefDiff.**

## 4.3 Evaluation of RefDiff by VIIRS/DNB

### *4.3.1 Reflectance of VIIRS/DNB*

Due to the lack of labels for evaluating the nighttime results generated by the model, we used data from the VNP02DNB product from VIIRS/DNB to assess model performance at night. To more accurately and objectively evaluate the model's precision at night, we first converted the VNP02DNB product into reflectance at the Top of Atmosphere (TOA):

$$R_{DNB} = \frac{\pi \cdot R_a}{I_{lunar,adj} \cdot cos\theta} \tag{11}$$

where $R_a$ is the radiance from the VNP02DNB product, with units of W/m²/sr/μm; $I_{lunar,adj}$ is the lunar irradiance, which must be adjusted based on the moon phase, Earth-Moon distance, and Earth-Sun distance (Miller and Turner, 2009), with units of W/m²/μm; and $\theta$ is the lunar zenith angle.

Due to the differences in the spectral response functions (SRF) between FY4B/AGRI and VIIRS/DNB, it is necessary to account for the impact of these discrepancies. Specifically, the wavelength ranges of the three visible light bands of AGRI are 0.45–0.49 μm, 0.55–0.75 μm, and 0.75–0.90 μm, while the wavelength range of the DNB is 0.5–0.9 μm. As shown in Figure 7, since the spectral range of DNB covers the spectral ranges of the AGRI bands centered at 0.65μm and 0.825μm, we need to adjust the AGRI reflectance to align with the DNB reflectance. Considering that the spectral range of the 0.47μm band lies outside the DNB range, it is not involved in the correction. The correction coefficient for band $i$ is given by $w_i$:

$$w_i = \frac{\int F_{AGRI,i}(\lambda) \cdot F_{DNB}(\lambda) \cdot I_{lunar,adj} d\lambda}{\int F_{DNB}(\lambda) \cdot I_{lunar,adj} d\lambda} \tag{12}$$

where $F_{AGRI,i}$ is the SRF of AGRI in band $i$, $F_{DNB}$ is the SRF of DNB, and $I_{lunar,adj}$ is the lunar irradiance spectrum (Miller and Turner, 2009).

For the computed $w_i$, we further adjust the reflectance generated by RefDiff to the spectral range of the DNB response function:

$$R_{AGRI,adj} = w_{0.65\mu m} \cdot R_{AGRI,0.65\mu m} + w_{0.825\mu m} \cdot R_{AGRI,0.825\mu m} \tag{13}$$

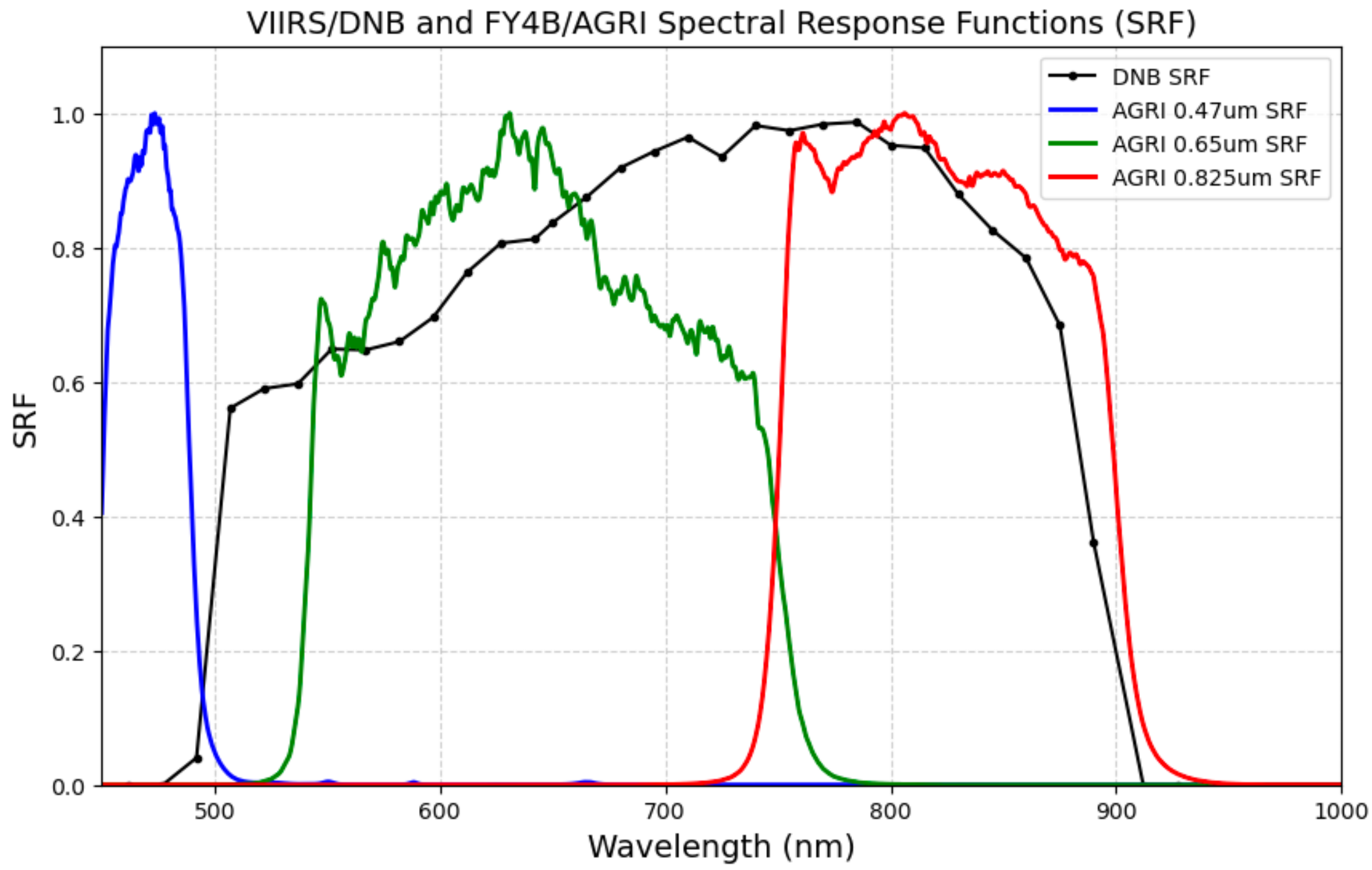

Figure 7. Spectral response functions of AGRI in three visible light bands and DNB.

#### *4.3.2 Evaluation of RefDiff performance at night by VIIRS/DNB*

After the above processing, we used $R_{DNB}$ to evaluate the performance of $R_{AGRI,adj}$. We analyzed 12 TC cases from June 2022 to December 2023, generating 336 cropped data samples for quantitative evaluation. Table 2 summarizes the generation results of the three models. As shown in Table 2, both the MAE and RMSE of the three models are higher at night than during the day. This may be due to the fact that the models were trained solely on daytime data, resulting in lower performance for nighttime predictions. Additionally, although the reflectance was corrected based on the sensor's SRF, the differences in sensor performance and observational angle may still influence the results, making complete correction difficult. Among the three models, RefDiff shows a slight increase in MAE at night (~0.048) compared to daytime (~0.034), while UNet's MAE increases from ~0.052 during the day to ~0.082 at night, and CGAN's MAE increases from ~0.051 to ~0.113. These results suggest that, in terms of MAE, RefDiff's performance remains comparable to its daytime results, whereas UNet and CGAN show a significant decrease in performance at night, particularly CGAN, which may be affected by domain shift, leading to a sharp decline in performance. In contrast to MAE, the increase in RMSE at night is more pronounced. This could be due to the large nighttime light values observed by VIIRS/DNB, which are not in the FY4B/AGRI data (since the AGRI is too far from the Earth to capture the light energy). Although this data constitutes a small proportion, its high values significantly impact the RMSE metric. For the SSIM metric, the nighttime performance of UNet and RefDiff shows little difference from daytime performance, while CGAN's nighttime performance is clearly lower than during the day. This further demonstrates that domain shift has a considerable impact on models like CGAN. For the PSNR metric, all three models show a notable increase at night compared to the day. This could be attributed to the fact that the numerator in the PSNR formula involves the maximum data value. As the nighttime light values from VIIRS/DNB are significantly high, their inclusion in the PSNR calculation leads to an increase in the PSNR values. In summary, although the performances of all models at night are affected by nighttime lighting and the distribution differences between daytime and nighttime data, RefDiff still exhibits the best performance, with the smallest discrepancy compared to the daytime results. This indicates that RefDiff possesses superior robustness and generalization ability.

**Table 2 Performance of the three models in nighttime mean results across three bands**

| Model | MAE | RMSE | SSIM | PSNR |
|---|---|---|---|---|
| UNet | 0.082 | 0.129 | 0.825 | 33.89 |
| CGAN | 0.113 | 0.164 | 0.766 | 31.81 |
| RefDiff | **0.048** | **0.090** | **0.880** | **37.32** |

We also visualized four nighttime TC cases, as shown in Figure 8. The specific times for the four cases are as follows: 18:15 UTC on September 12, 2022, 16:15 UTC on June 1, 2023, 17:00 UTC on August 29, 2023, and 18:00 UTC on October 2, 2023. In the Figure 8, the four columns on the left show the DNB reflectance and the results generated by each model, while the three columns on the right display the differences between the model results and the DNB reflectance. From the figure, it is evident that the DNB reflectance data clearly reveals the structure and texture of the TC cases. While UNet captures the general outlines of the TC, it fails to effectively reconstruct the detailed structures. CGAN provides a more accurate representation of the textures of the TC, but also introduces some erroneous data (Figure 9), which explains the increase in MAE and RMSE for CGAN at night compared to during the day. In contrast, RefDiff maintains the best overall performance, being less affected by such disturbances. These comparisons highlight RefDiff's distinct advantage in visible light generation tasks. The trends shown in the three rightmost columns further confirm this pattern. RefDiff has the smallest difference compared to the DNB reflectance, followed by UNet, while CGAN exhibits the largest discrepancy. Moreover, we observed that in areas with clouds, both UNet and CGAN results, although below the standard values during the day, are consistently above the standard values at night.

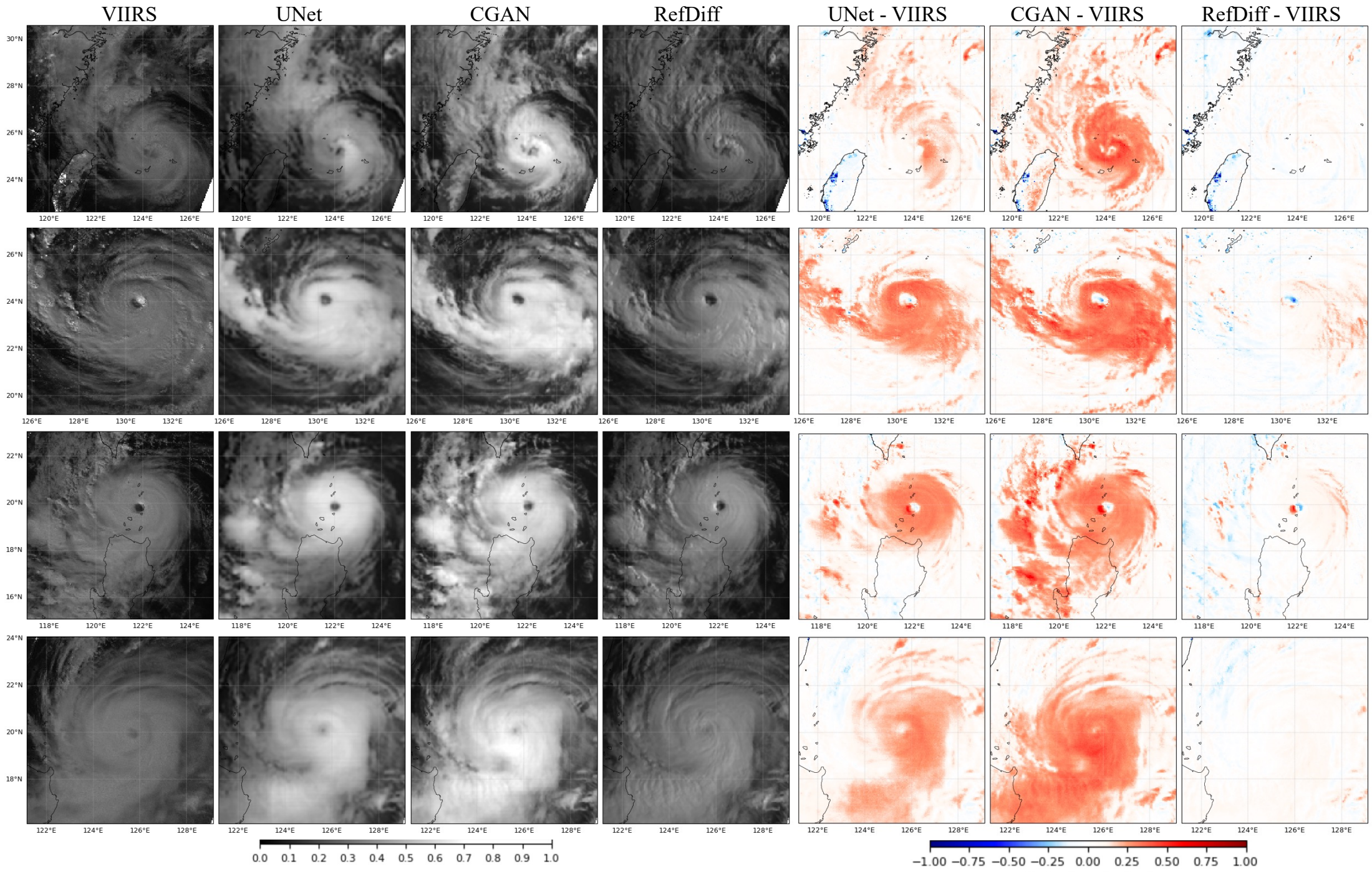

**Figure 8. Reflectance of DNB and results of the three models across various TC cases. The first column presents the reflectance from the VIIRS product. The second to the fourth columns display the results from the UNet, CGAN, and RefDiff models, respectively. The fifth to the seventh columns show the differences between the model results and the VIIRS reflectance.**

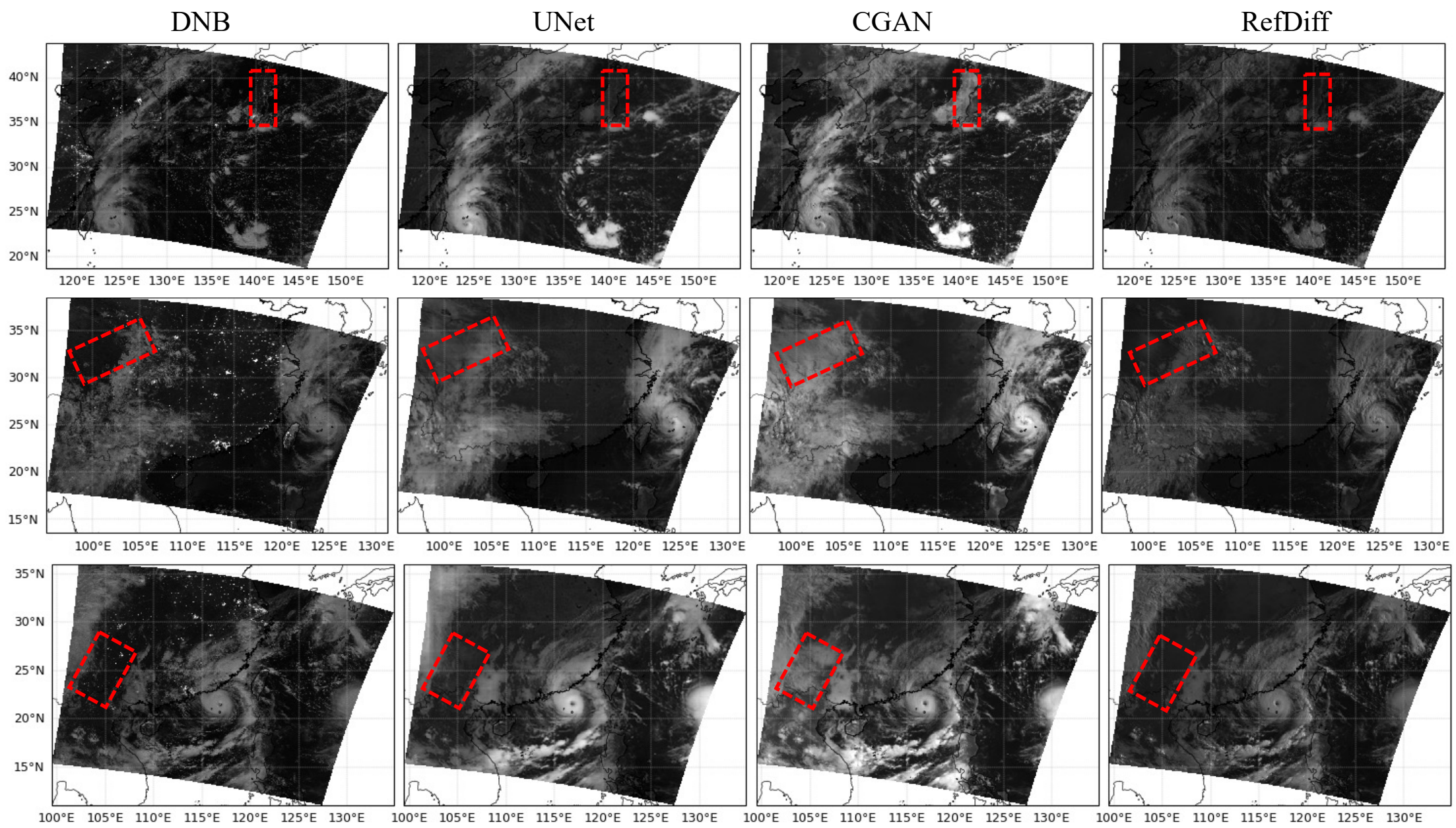

**Figure 9. The impact of domain shift on the CGAN model, where the red-boxed area shows clouds generated by the CGAN model that do not actually exist.**

## 5. Conclusion

This study applies a generative diffusion model, named RefDiff, to develop a high-precision nighttime visible light generation model based on multi-band brightness temperature data from the FY4B/AGRI sensor, along with other auxiliary data. RefDiff enables simultaneously generation of visible light data across the 0.47 μm, 0.65 μm, and 0.825 μm bands.

As a probabilistic model, RefDiff not only improves the model's accuracy through multiple iterations and ensemble mean, but also provides uncertainty to reflect the model's reliability. In terms of performance, RefDiff significantly surpasses traditional models such as UNet and CGAN. For multi-band visible light generation, the mean MAE decreased from 0.0517 to 0.0337, while SSIM increased from 0.801 to 0.904, showcasing exceptional performance in regions with complex, thick cloud cover. We also calculated the nighttime reflectance using the VIIRS product to evaluate the model's performance on nighttime results. The results indicate that RefDiff performs comparably to its daytime performance. This advancement broadens the potential applications of visible light data during the night, particularly in critical meteorological areas, such as real-time, all-day continuous monitoring of TC and severe convective systems.

## Acknowledgment

This research was supported by the National Key Research and Development Program of China (Grant (2024YFF0808303), and was also supported by Zhejiang Provincial Natural Science Foundation of China under Grant No.LQN25D050004. The authors sincerely acknowledge the Fengyun Satellite Data Center in China and National Aeronautics and Space Administration in American for providing Fengyun-4B products, and VIIRS Panchromatic Day-Night band (DNB) Calibrated Radiance Product. In addition, they also acknowledge the European Centre for Medium-Range Weather Forecasts (ECMWF) to provide Land Cover Classification Gridded Maps.

**Author contributions:** Conceptualization: T.Z., F.Z., H.F. Methodology: T.Z., F.Z., H.X., B.P. Investigation: T.Z., F.Z., Visualization: T.Z. Supervision: T.Z., F.Z., H.F. Writing—original draft: T.Z., H.F. Writing—review & editing: B.P., Z.Y., F.L.

**Competing interests:** The authors declare that they have no competing interests.

## References

American Meteorological Society, 2009. Bulletin of the American Meteorological Society - American Meteorological Society - Google Books. American Meteorological Society.

Bessho, K., Date, K., Hayashi, M., Ikeda, A., Imai, T., Inoue, H., Kumagai, Y., Miyakawa, T., Murata, H., Ohno, T., Okuyama, A., Oyama, R., Sasaki, Y., Shimazu, Y., Shimoji, K., Sumida, Y., Suzuki, M., Taniguchi, H., Tsuchiyama, H., Uesawa, D., Yokota, H., Yoshida, R., 2016. An Introduction to Himawari-8/9— Japan’s New-Generation Geostationary Meteorological Satellites. Journal of the Meteorological Society of Japan 94, 151–183.

https://doi.org/10.2151/jmsj.2016-009
Brunet, D., Vrscay, E.R., Zhou Wang, 2012. On the Mathematical Properties of the Structural Similarity Index. IEEE Trans. on Image Process. 21, 1488–1499. https://doi.org/10.1109/TIP.2011.2173206
Chen, Z., Yu, B., Yang, C., Zhou, Y., Yao, S., Qian, X., Wang, C., Wu, B., Wu, J., 2021. An extended time series (2000–2018) of global NPP-VIIRS-like nighttime light data from a cross-sensor calibration. Earth System Science Data 13, 889–906. https://doi.org/10.5194/essd-13-889-2021
Chollet, F., 2016. Xception: Deep Learning with Depthwise Separable Convolutions. https://doi.org/10.48550/ARXIV.1610.02357
Cordonnier, J.-B., Loukas, A., Jaggi, M., 2021. Multi-Head Attention: Collaborate Instead of Concatenate. https://doi.org/10.48550/arXiv.2006.16362
Dhariwal, P., Nichol, A., 2021. Diffusion Models Beat GANs on Image Synthesis, in: Advances in Neural Information Processing Systems. Curran Associates, Inc., pp. 8780–8794.
Elvidge, C.D., Baugh, K., Zhizhin, M., Hsu, F.C., Ghosh, T., 2017. VIIRS night-time lights. International Journal of Remote Sensing 38, 5860–5879. https://doi.org/10.1080/01431161.2017.1342050
Gao, Z., Shi, X., Han, B., Wang, H., Jin, X., Maddix, D., Zhu, Y., Li, M., Wang, Y., 2023. PreDiff: Precipitation Nowcasting with Latent Diffusion Models. https://doi.org/10.48550/arXiv.2307.10422
Goodfellow, I., Pouget-Abadie, J., Mirza, M., Xu, B., Warde-Farley, D., Ozair, S., Courville, A., Bengio, Y., 2020. Generative adversarial networks. Commun. ACM 63, 139–144. https://doi.org/10.1145/3422622
Goodman, S.J., Schmit, T.J., Daniels, J., Redmon, R.J., 2019. The GOES-R Series: A New Generation of Geostationary Environmental Satellites. Elsevier.
Han, K.-H., Jang, J.-C., Ryu, S., Sohn, E.-H., Hong, S., 2022. Hypothetical Visible Bands of Advanced Meteorological Imager Onboard the Geostationary Korea Multi-Purpose Satellite -2A Using Data-To-Data Translation. IEEE Journal of Selected Topics in Applied Earth Observations and Remote Sensing 15, 8378–8388. https://doi.org/10.1109/JSTARS.2022.3210143
Ho, J., Jain, A., Abbeel, P., 2020. Denoising Diffusion Probabilistic Models, in: Advances in Neural Information Processing Systems. Curran Associates, Inc., pp. 6840–6851.
Ho, J., Salimans, T., 2022. Classifier-Free Diffusion Guidance. https://doi.org/10.48550/arXiv.2207.12598
Huang, L., Gianinazzi, L., Yu, Y., Dueben, P.D., Hoefler, T., 2024. DiffDA: a Diffusion Model for Weather-scale Data Assimilation. https://doi.org/10.48550/arXiv.2401.05932
Kim, J.-H., Ryu, S., Jeong, J., So, D., Ban, H.-J., Hong, S., 2020. Impact of Satellite Sounding Data on Virtual Visible Imagery Generation Using Conditional Generative Adversarial Network. IEEE Journal of Selected Topics in Applied Earth Observations and Remote Sensing 13, 4532–4541. https://doi.org/10.1109/JSTARS.2020.3013598
Kim, K., Kim, J.-H., Moon, Y.-J., Park, E., Shin, G., Kim, T., Kim, Y., Hong, S., 2019. Nighttime Reflectance Generation in the Visible Band of Satellites. Remote Sensing 11, 2087. https://doi.org/10.3390/rs11182087
Korhonen, J., You, J., 2012. Peak signal-to-noise ratio revisited: Is simple beautiful?, in: 2012 Fourth

International Workshop on Quality of Multimedia Experience. Presented at the 2012 Fourth International Workshop on Quality of Multimedia Experience (QoMEX 2012), IEEE, Melbourne, Australia, pp. 37–38. https://doi.org/10.1109/QoMEX.2012.6263880

Li, L., Carver, R., Lopez-Gomez, I., Sha, F., Anderson, J., 2024. Generative emulation of weather forecast ensembles with diffusion models. Science Advances 10, eadk4489. https://doi.org/10.1126/sciadv.adk4489

Miller, S.D., Turner, R.E., 2009. A Dynamic Lunar Spectral Irradiance Data Set for NPOESS/VIIRS Day/Night Band Nighttime Environmental Applications. IEEE Trans. Geosci. Remote Sensing 47, 2316–2329. https://doi.org/10.1109/TGRS.2009.2012696

Min, M., Li, J., Wang, F., Liu, Z., Menzel, W.P., 2020. Retrieval of cloud top properties from advanced geostationary satellite imager measurements based on machine learning algorithms. Remote Sensing of Environment 239, 111616. https://doi.org/10.1016/j.rse.2019.111616

Mirza, M., Osindero, S., 2014. Conditional Generative Adversarial Nets. https://doi.org/10.48550/arXiv.1411.1784

Rombach, R., Blattmann, A., Lorenz, D., Esser, P., Ommer, B., 2022. High-Resolution Image Synthesis With Latent Diffusion Models. Presented at the Proceedings of the IEEE/CVF Conference on Computer Vision and Pattern Recognition, pp. 10684–10695.

Ronneberger, O., Fischer, P., Brox, T., 2015. U-Net: Convolutional Networks for Biomedical Image Segmentation. https://doi.org/10.48550/arXiv.1505.04597

Schmit, T.J., Griffith, P., Gunshor, M.M., Daniels, J.M., Goodman, S.J., Lebair, W.J., 2017. A Closer Look at the ABI on the GOES-R Series. Bulletin of the American Meteorological Society 98, 681–698. https://doi.org/10.1175/BAMS-D-15-00230.1

Stuhlmann, R., Rodriguez, A., Tjemkes, S., Grandell, J., Arriaga, A., Bézy, J.-L., Aminou, D., Bensi, P., 2005. Plans for EUMETSAT's Third Generation Meteosat geostationary satellite programme. Advances in Space Research, Atmospheric Remote Sensing: Earth's Surface, Troposphere, Stratosphere and Mesosphere- I 36, 975–981. https://doi.org/10.1016/j.asr.2005.03.091

Task, H.L., 2001. Night vision goggle visual acuity assessment: results of an interagency test, in: Lewandowski, R.J., Haworth, L.A., Girolamo, H.J., Rash, C.E. (Eds.), . Presented at the Aerospace/Defense Sensing, Simulation, and Controls, Orlando, FL, pp. 130–137. https://doi.org/10.1117/12.437987

Velden, C., Harper, B., Wells, F., Beven, J.L., Zehr, R., Olander, T., Mayfield, M., Guard, C. “Chip,” Lander, M., Edson, R., Avila, L., Burton, A., Turk, M., Kikuchi, A., Christian, A., Caroff, P., McCrone, P., 2006. The Dvorak Tropical Cyclone Intensity Estimation Technique: A Satellite-Based Method that Has Endured for over 30 Years. Bull. Amer. Meteor. Soc. 87, 1195–1210. https://doi.org/10.1175/BAMS-87-9-1195

Wang, Z., Bovik, A.C., 2006. Modern Image Quality Assessment, Synthesis Lectures on Image, Video, and Multimedia Processing. Springer International Publishing, Cham. https://doi.org/10.1007/978-3-031-02238-8

Wang, Z., Bovik, A.C., Sheikh, H.R., Simoncelli, E.P., 2004. Image Quality Assessment: From Error Visibility to Structural Similarity. IEEE Trans. on Image Process. 13, 600–612. https://doi.org/10.1109/TIP.2003.819861

Wang, Z., Simoncelli, E.P., Bovik, A.C., 2003. Multiscale structural similarity for image quality assessment, in: The Thrity-Seventh Asilomar Conference on Signals, Systems & Computers, 2003. Presented at the Conference Record of the 37th Asilomar Conference on Signals, Systems and Computers, IEEE, Pacific Grove, CA, USA, pp. 1398–1402. https://doi.org/10.1109/ACSSC.2003.1292216

Xian, D., Zhang, P., Gao, L., Sun, R., Zhang, H., Jia, X., 2021. Fengyun Meteorological Satellite Products for Earth System Science Applications. Adv. Atmos. Sci. 38, 1267–1284. https://doi.org/10.1007/s00376-021-0425-3

Xiao, H., Zhang, F., Wang, L., Li, W., Guo, B., Li, J., 2024. CloudDiff: Super-resolution ensemble retrieval of cloud properties for all day using the generative diffusion model. https://doi.org/10.48550/arXiv.2405.04483

Xiao, Y., Yuan, Q., Jiang, K., He, J., Jin, X., Zhang, L., 2024. EDiffSR: An Efficient Diffusion Probabilistic Model for Remote Sensing Image Super-Resolution. IEEE Transactions on Geoscience and Remote Sensing 62, 1–14. https://doi.org/10.1109/TGRS.2023.3341437

Yao, J., Du, P., Zhao, Y., Wang, Y., 2024. Simulating Nighttime Visible Satellite Imagery of Tropical Cyclones Using Conditional Generative Adversarial Networks.

Yuan, Q., Shen, H., Li, T., Li, Z., Li, S., Jiang, Y., Xu, H., Tan, W., Yang, Q., Wang, J., Gao, J., Zhang, L., 2020. Deep learning in environmental remote sensing: Achievements and challenges. Remote Sensing of Environment 241, 111716. https://doi.org/10.1016/j.rse.2020.111716

Zhang, H., Zhang, Z., Odena, A., Lee, H., 2019. Consistency Regularization for Generative Adversarial Networks. https://doi.org/10.48550/ARXIV.1910.12027

Zhang, K., Sun, M., Han, T.X., Yuan, X., Guo, L., Liu, T., 2018. Residual Networks of Residual Networks: Multilevel Residual Networks. IEEE Trans. Circuits Syst. Video Technol. 28, 1303–1314. https://doi.org/10.1109/TCSVT.2017.2654543

Zhang, Q., Tao, M., Chen, Y., 2023. gDDIM: Generalized denoising diffusion implicit models. https://doi.org/10.48550/arXiv.2206.05564

Zhao, Z., Zhang, F., Wu, Q., Li, Z., Tong, X., Li, J., Han, W., 2023. Cloud Identification and Properties Retrieval of the Fengyun-4A Satellite Using a ResUnet Model. IEEE Trans. Geosci. Remote Sensing 61, 1–18. https://doi.org/10.1109/TGRS.2023.3252023

Zhou, Z., Siddiquee, M.M.R., Tajbakhsh, N., Liang, J., 2018. UNet++: A Nested U-Net Architecture for Medical Image Segmentation. https://doi.org/10.48550/ARXIV.1807.10165